\documentclass[10pt,twocolumn,letterpaper]{article}

\usepackage{cvpr}              % To produce the CAMERA-READY version
\definecolor{cvprblue}{rgb}{0.21,0.49,0.74}
\usepackage[pagebackref,breaklinks,colorlinks,allcolors=cvprblue]{hyperref}
\usepackage{float}

\def\paperID{5615} % *** Enter the Paper ID here
\def\confName{CVPR}
\def\confYear{2025}

\title{FilmComposer: LLM-Driven Music Production for Silent Film Clips
}

\author{Zhifeng Xie$^{1,2}$, Qile He$^{1}$, Youjia Zhu$^{1}$, Qiwei He$^{1}$, Mengtian Li$^{1,2}$\footnotemark[2]\\
$^{1}$Shanghai University\\
$^{2}$Shanghai Engineering Research Center of Motion Picture Special Effects\\
{\tt\small \{zhifeng\_xie,shu\_hql,zyj028,heqiwei,mtli\}@shu.edu.cn}}
\begin{document}
\maketitle
\renewcommand{\thefootnote}{\fnsymbol{footnote}}
\footnotetext[2]{Corresponding author.}
\begin{abstract}
In this work, we implement music production for silent film clips using LLM-driven method. 
Given the strong professional demands of film music production, we propose the \textbf{FilmComposer}, simulating the actual workflows of professional musicians. 
FilmComposer is the first to combine large generative models with a multi-agent approach, leveraging the advantages of both waveform music and symbolic music generation. Additionally, FilmComposer is the first to focus on the three core elements of music production for film—audio quality, musicality, and musical development—and introduces various controls, such as rhythm, semantics, and visuals, to enhance these key aspects. 
Specifically, FilmComposer consists of the visual processing module, rhythm-controllable MusicGen, and multi-agent assessment, arrangement and mix. In addition, our framework can seamlessly integrate into the actual music production pipeline and allows user intervention in every step, providing strong interactivity and a high degree of creative freedom. 
Furthermore, we propose \textbf{MusicPro-7k} which includes 7,418 film clips, music, description, rhythm spots and main melody, considering the lack of a professional and high-quality film music dataset. 
Finally, both the standard metrics and the new specialized metrics we propose demonstrate that the music generated by our model achieves state-of-the-art performance in terms of quality, consistency with video, diversity, musicality, and musical development. 
Project page: \url{https://apple-jun.github.io/FilmComposer.github.io/}
\end{abstract} 
\section{Introduction}
\label{sec:intro}
The music in films can enhance visual content, guide emotions, and support narrative flow. However, manual music production is time-consuming and requires specialized expertise. Therefore, developing a fast, effective and user-friendly AI approach for film music production would assist experts in producing music more efficiently, while also enabling non-experts to create high-quality music with ease.

The challenge of AI music production lies in meeting the high standards of quality, musicality, and development. Typical music for film demands at least a 48kHz sampling rate and 24-bit depth, which current models are still far away from. It requires not only technical excellence but also heightened musicality, encompassing both expression and aesthetics. The most overlooked one, Musical development, refers to the evolution of themes and motifs over time, greatly contributing to audio-visual consistency.

%-----------------------------------------------------------------------
\begin{figure}[t]
  \centering
   \includegraphics[width=0.48\textwidth]{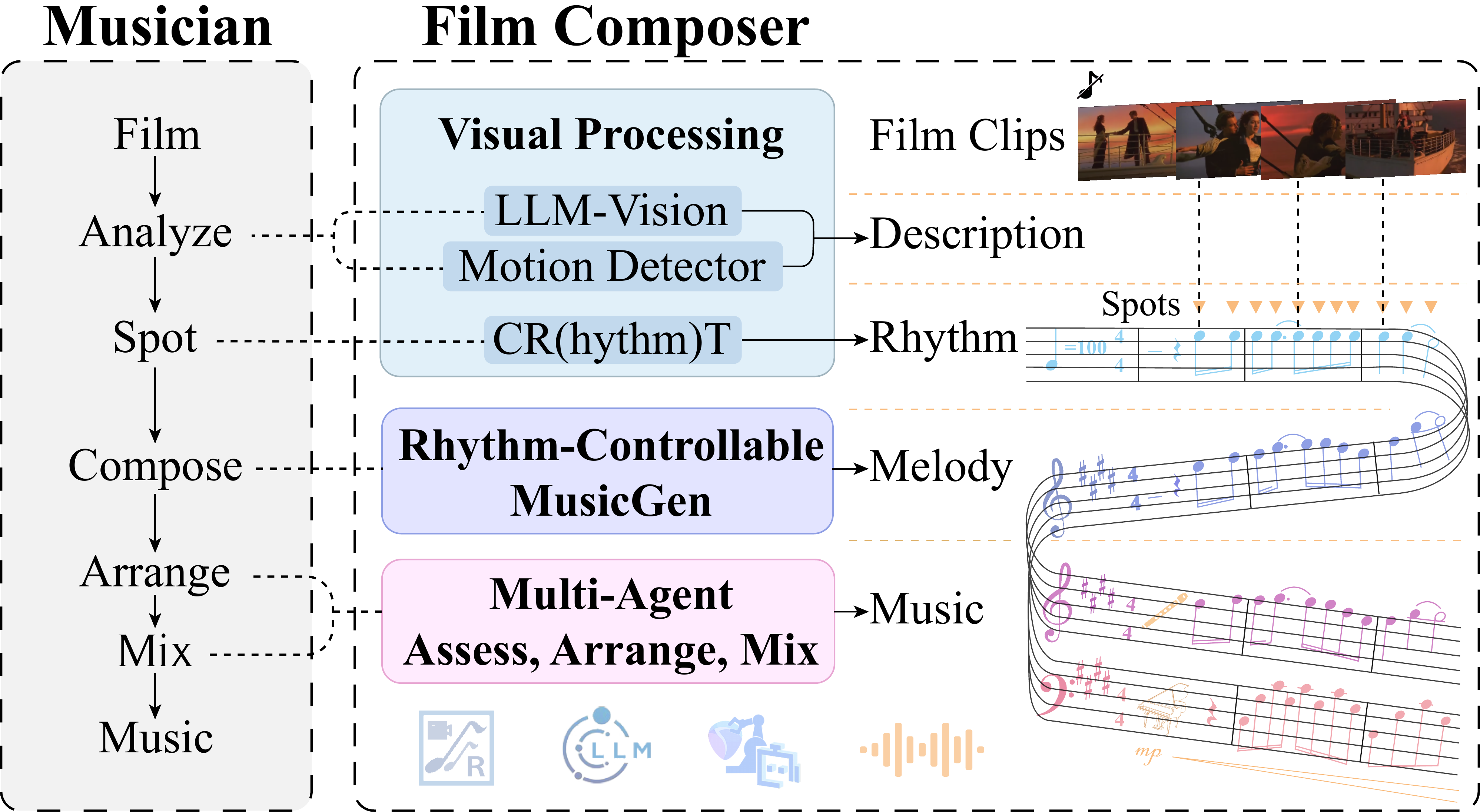}
   \setlength{\abovecaptionskip}{-2mm}
   \caption{\textbf{A schematic of our work.} The left column illustrates the actual steps taken by human musicians in music production, while the middle column represents the corresponding simulated blocks in the FilmComposer. The inputs and outputs at each stage are depicted on the right, visually demonstrating how film clips are gradually transformed into the final music. }
   \vspace{-6mm}
   \label{fig:intro}
\end{figure}
%-----------------------------------------------------------------------

Previous work focused on optimizing audio token compression to improve music quality. However, even the most advanced token interleaving patterns~\cite{copet2024simple} and stable audio techniques~\cite{evans2024fast} fail to meet the required standard. Since the introduction of MusicLM~\cite{agostinelli2023musiclm}, large language models have gained significant attention, notably improving musicality. Following the open-sourcing of Meta's MusicGen~\cite{copet2024simple}, large language models have been widely integrated into various frameworks \cite{liu2023m,lan2024musicongen}. While musicality has improved, control remains limited. Some research has explored adding conditions, such as rhythm \cite{di2021video,zhuo2023video,lan2024musicongen,li2024diff}, timing \cite{evans2024fast}, and melody \cite{copet2024simple,lan2024musicongen}. Yet, these methods tend to either compromise musicality or only achieve superficial control, lacking effective musical development. In summary, breakthroughs have been made, but it remains insufficient for the specific demands of music production for film.

To address the challenges above, we propose \textbf{FilmComposer} which imitates musicians to produce music for film, including spotting, composition, arrangement and mix. Corresponding to the general three steps, FilmComposer consists of visual processing, rhythm-controllable MusicGen, and multi-agent assessment, arrangement and mix. To train FilmComposer, We also construct a professional film music dataset named \textbf{MusicPro-7k}, comprising 7K video-music pairs, descriptions, main melodies, and rhythm spots.

Specifically, we process film clips to build description and precise rhythm control for spotting. Next, we integrate a rhythm conditioner into MusicGen and fine-tune it to compose in sync with rhythm cues, ensuring strong musicality and video alignment. This adaptation makes the model the first large language model for film-to-music generation, with generating music from visual input lowering the threshold of large language model usage. Additionally, we designed an agent-based musicality metric to filter generated melodies, ensuring a main melody with strong musicality. Next, we develop a multi-agent system based on large language models to arrange the generated main melody, which means organizing musical elements to support thematic, emotional and narrative progression. It enhances the musicality and provides excellent musical development. Then the multi-agent system operates a digital audio workstation (DAW) to arrange, mix, and finally output a high-quality and ready-to-use music for silent film clips. Our contributions are summarized as follows: 
\begin{itemize}[leftmargin=1em] 
    \item We propose the \textbf{FilmComposer}, functioning as a musician, highly interactive and seamlessly integrated into actual production pipelines. 
    
    \item We construct Rhythm-Controllable MusicGen and fine-tune it by self-created multifunctional film music dataset \textbf{MusicPro-7k}, enabling it to generate music that aligns with film clips while maintaining strong musicality. 
    
    \item We integrate a multi-agent system to enhance musical development, resulting in music that excels in quality, diversity, musicality, and audio-visual consistency, as validated by newly proposed evaluations and user study.
\end{itemize}
%-------------------------------------------------------------------------
\begin{figure*}
    \centering
    \includegraphics[width=1\textwidth]{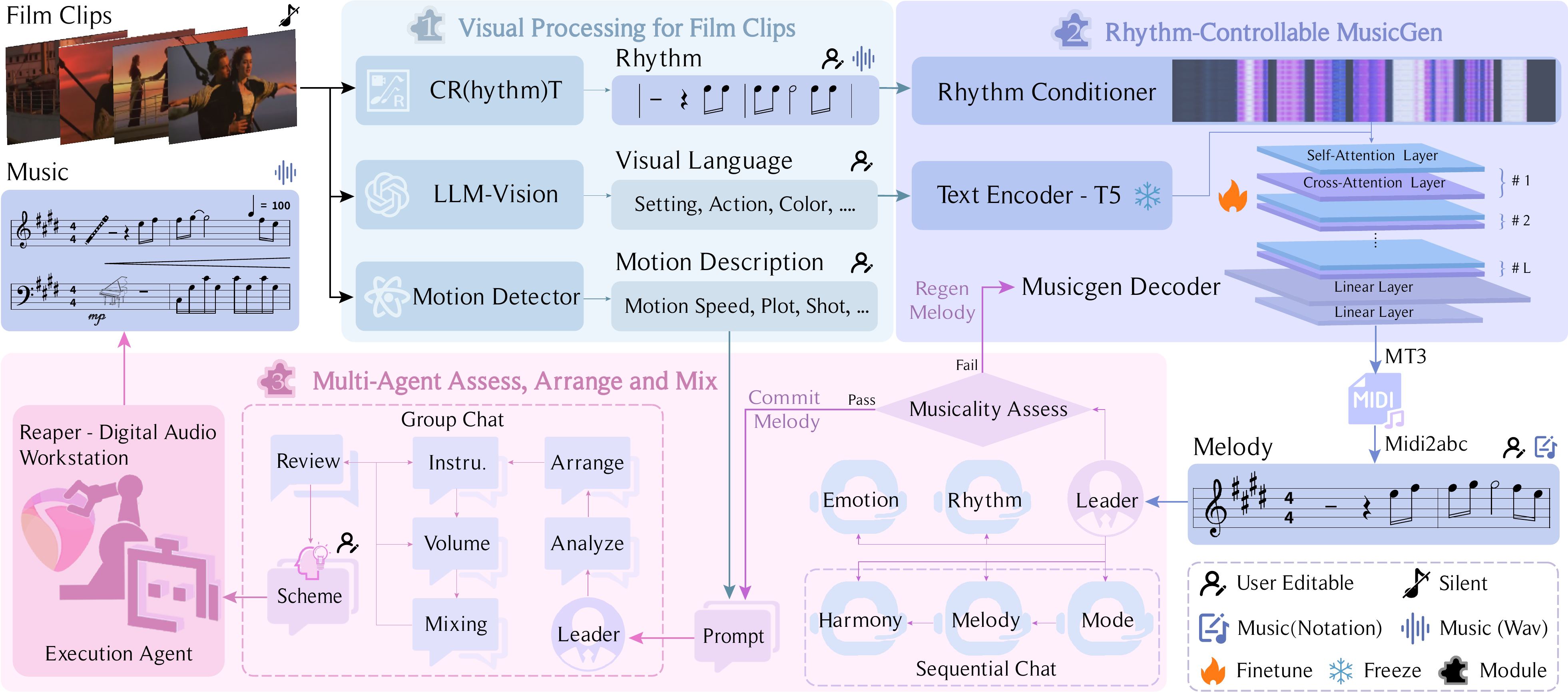}
    \caption{\textbf{The framework of FilmComposer.} Three large color blocks represent the three main modules, through which the input Film clips pass sequentially, ultimately outputting a waveform. The three blue blocks with musical notation illustrate the complete music production process, from setting the rhythm points, composing, to arranging and mixing.}
    \vspace{-4mm}
    \label{fig:pipeline}
\end{figure*}

\section{Related Work}
\label{sec:related}
\subsection{Music Generation}
\label{sec:music-generation}
\noindent \textbf{Waveform music generation.}
The question of whether to use symbolic music generation or waveform music generation remains a contentious issue. Waveform music benefits from its prevalence, supported by a wealth of generation models \cite{agostinelli2023musiclm,huang2023noise2music,schneider2024mousai,copet2024simple,liu2024audioldm,evans2024fast,lan2024musicongen} and vast amounts of music data \cite{agostinelli2023musiclm,huang2023noise2music,schneider2024mousai,evans2024fast}, resulting in typically richer outputs. However, it suffers from generally poor sound quality, frequent noise and distortion, and less effective control.

\noindent \textbf{Symbolic music generation.}
Symbolic music records how music is performed and thus captures the essence of music more closely, allowing for finer control and resulting in more pleasing melodies \cite{lee2022commu,yu2023musicagent,deng2024composerx,cheng2024autostudio}. Leveraging the unique characteristics of symbolic music, some studies have moved away from commonly used models like transformers and diffusion for music generation. Instead, they employ the multi-agent approach, achieving impressive results with greater simplicity \cite{yu2023musicagent,deng2024composerx,cheng2024autostudio}. However, symbolic music generation relies on MIDI datasets, which are scarce and resource-intensive to build. As a result, symbolic music generation models struggle to achieve large-scale deployment, leading to repetitive and monotonous music. 

Aiming to harness the richness of waveform music while leveraging the high quality of symbolic music, we propose FilmComposer, which effectively integrates both strategies.
%-------------------------------------------------------------------------
\subsection{Video to Music Task}
\label{sec:video-to-music}
\noindent \textbf{Video to music generation.}
The task of generating music for human-centric videos has been around for some time, primarily focusing on dance and concert \cite{yu2023long,zhu2022quantized,gan2020foley}, where clear human movements dictate the musical rhythm. Subsequently, CMT \cite{di2021video} pioneered the concept of video background music generation, enabling music creation for more flexible video formats, particularly suitable for short videos. CMT focuses on the rhythm alignment, connecting video and music by extracting their corresponding features. Diff-bgm \cite{li2024diff} continued to explore rhythm control, yielding new advancements. Additionally, certain studies \cite{zhuo2023video,kang2024video2music,liu2023m} have focused on semantic alignment, with M2UGen \cite{liu2023m} incorporating text controls for music generation, achieving impressive results. VidMuse \cite{tian2024vidmuse} has begun exploring more specialized scenarios, such as music generation for advertisements and trailers, employing long-short sequence modeling to achieve multiple controls simultaneously. 

\noindent \textbf{Music retrieval.}
Music Retrieval also receives attention in recent studies \cite{pretet2021cross,suris2022s,yi2021cross,zeng2018audio}, with some works examining the relationship between video, audio and language \cite{guzhov2022audioclip,wu2022wav2clip}, others guiding music recommendations based on description \cite{mckee2023language} or dialogue \cite{dong2024musechat}. 

However, the works above ignored exploring film music or considering the actual requirement of music production. Therefore, our work is the first to focus specifically on music production for film, requiring the generated music to be more professional and closely aligned with various aspects of film clips. Meanwhile, our approach integrates more user-friendly rhythm control and text control, resulting in more refined results.

%-------------------------------------------------------------------------
\subsection{Dataset}
\label{sec:dataset_related_work}
Most video-to-music datasets only focus on generating music for human-centric videos---dance \cite{zhu2022quantized,li2021ai,yi2021cross} or concert \cite{liu2023m}. MuVi-Sync \cite{kang2024video2music} collects music videos characterized by a pop-oriented style, unsuitable for film music production. V2M \cite{tian2024vidmuse} has gathered a more diverse set of video music data, but it primarily covers advertisements and trailers, with severe audio degradation caused by the mute operations. It leans toward advertising and differs from true film music. Since the professional requirements for music production are significantly higher than these types of music, there is still a lack of high-quality, large-scale, specialized film music datasets. To address this problem, we construct a dedicated dataset MusicPro-7k including film clips and music pairs, descriptions and rhythm spots.
\section{Method}
\label{sec:method}

% This section introduces the FilmComposer framework, 
\noindent \textbf{Overview.} As shown in \cref{fig:pipeline}, FilmComposer consists of three main modules---visual processing, rhythm-controllable MusicGen and multi-agent assessment, arrangement and mix---to simulate the process of a human musician producing music for film. In \cref{sec:processing}, we perform visual processing on input film clips to construct rhythm conditions, shape visual language, and extract motion descriptions, simulating the analysis and spotting process. In \cref{sec:musicgen}, the Rhythm-Controllable MusicGen generates melodies with high musicality, aligned with the rhythm spots and visual language. The melody is transcribed into MIDI and ABC notation for further professional work. In \cref{sec:agent}, a multi-agent system assesses the melody's musicality. if it fails to meet the standard, the previous module will be ordered to generate a new melody. Otherwise, the system proceeds with arrangement and mix, mapping out a scheme and instructing a digital audio workstation to execute it, ultimately generating cinematic-quality music. 
%------------------------------------------------------------------------
\subsection{Visual Processing for Film Clips}
\label{sec:processing}

When generating music for film clips, three key aspects must be considered: first, the music should align with the visual content to enhance the atmosphere; second, it must follow the pacing of the film clips; and third, it should reflect higher-level film semantics, such as emotions, themes, and plot development. To address this, we do visual processing to extract these three components from the film clips.

First, we modify CMT \cite{di2021video} into the CRT---Controllable Rhythm Transformer, which processes video and generates rhythm points. To be specific, we introduce an algorithm that extracts the main melody from symbolic music, and then flatten it into a rhythm. The algorithm calculates track coverage and note ratios, using instrument performance characteristics to identify the lead instrument. The coverage $Coverage_i$ for the $i$-th track is:
\begin{equation} 
    Coverage_i = \frac{T_{i}}{T_{music}}, 
    \label{eq1} 
\end{equation}
where $T_{i}$ is the total time during which the notes of the $i$-th track are played, and $T_{music}$ is the music duration.

The note ratio $R_i$ for the $i$-th track is calculated as:
\begin{equation}
  R_i = \frac{n_i}{\sum_{i=1}^{n} n_i},
  \label{eq:note-ratio-combined}
\end{equation}
where $n_i$ is the number of notes in the $i$-th track.

If the lead instrument’s coverage is insufficient, we incorporate additional tracks until the total coverage $Coverage$ meets the threshold, computed as:
\begin{equation}
     Coverage = \frac{1}{T_{music}} \sum_{i=1}^{2n-1} \mathbb{I}(N(t_i) > 0) \cdot (t_{i+1} - t_i),
    \label{eq:coverage}
\end{equation}
where $T_{music}$ is total music duration; $\{t_i\}_{i=1}^{2n}$ is ascending sequence of all note-onset and note-offset timestamps; $N(t)$ is active note count at $t$, incremented for note-onset and decremented for note-offset; $\mathbb{I}(\cdot)$ is indicator function.

Second, drawing from audio-visual language and music emotion analysis theory \cite{han2022survey, hevner1935affective,russell1980circumplex}, we designed visual attributes: setting, brightness, color hue, action, emotional tone, view scale, and theme. ChatGPT-4V \cite{achiam2023gpt} is used to efficiently recognize these attributes from selected frames.

The elements extracted above are enough to guide the composition, but more is needed to guide the arrangement and mix. So third, We employ open-source algorithms \cite{bradski2000opencv,SceneDetect,achiam2023gpt} to extract motion speed, motion saliency, shot cut, and plot development. Motion speed $S_{motion}$ for each video segment is calculated as:
\begin{equation} 
    S_{motion} = \frac{1}{N} \sum_{i=1}^{N} M_i, 
    \label{eq4} 
\end{equation}
where $M_i$ is the optical flow magnitude at the $i$-th frame. 

The motion saliency $S_{saliency}$ is the mean of the positive changes in optical flow magnitude between consecutive frames---the average increase in motion intensity:
\begin{equation} 
    S_{saliency} = \frac{1}{N^+} \sum_{i=1}^{N^+} (M_i - M_{i-1})_+. 
    \label{eq5} 
\end{equation}

%------------------------------------------------------------------------
\subsection{Rhythm-Controllable MusicGen}
\label{sec:musicgen}

Once rhythm spots and semantic information are ready, large language models can be applied to generate the corresponding main melody. Since there is no music generation model that can take visual language as input and be controlled by rhythm spots, we construct Rhythm-Controllable MusicGen and train it with MusicPro-7k.

Rhythm-Controllable MusicGen is composed of the rhythm conditioner, the T5 text encoder \cite{raffel2020exploring}, and the MusicGen decoder \cite{copet2024simple}. In general, we explore joint conditioning of both the chromagram of the input rhythm spots and visual description to exert control over the melodic structure. The output $Y_{output}$ can be calculated as:
\begin{equation} 
    Y_{output} = \text{Decoder}(C_{rhythm}, C_{description}, Y),
    \label{eq_rc_musicgen} 
\end{equation}
where the rhythm condition $C_{rhythm}$ and visual description $C_{description}$ is passed as prefix to the transformer input $Y$ - codebook projections along with the positional embeddings. specifically, we employ the prepend method for condition fuser, sequentially concatenating the processed rhythm condition, text condition, and transformer input. 

The input passes through a transformer consisting of $L$ layers, each with dimension $D$. Every layer incorporates a causal self-attention block and a cross-attention block. Lastly, the layer concludes with a fully connected block composed of a linear layer from $D$ to 4·$D$ channels, followed by a ReLU activation, and then another linear layer back to $D$. 

We focus on implementing the rhythm conditioner that processes audio inputs for model conditioning. This specific conditioner operates based on the chromagram, which reflects the detailed beat characteristics of the audio. The input waveform of rhythm spots is tokenized and the chromagram features are extracted from it. These chromagram embeddings are projected to match the required output dimensions. A downsampling factor is applied to adjust the chromagram length, and a mask ensures that only valid audio regions are considered. The system supports both real-time chromagram extraction and precomputed chromagrams, enhancing flexibility in handling different audio inputs.

%------------------------------------------------------------------------
\subsection{Multi-Agent Assess, Arrange and Mix}
\label{sec:agent}
To make the following operations efficient, we use the MT3 model \cite{gardner2021mt3} to convert the music to the MIDI format and then use Midi2abc \cite{midi2abc} to convert it into ABC notation-a musical notation format based on the ASCII character set, aligned with the musician's composition process.

To ensure the musicality of the melody generated, we design a multi-agent assessment system based on AutoGen framework \cite{wu2023autogen}. If the melody meets the criteria, it proceeds to the arrangement part; otherwise, the rhythm-controllable MusicGen module regenerates it. Each agent acts as a reviewer, guided by Role-play prompt engineering, evaluating aspects such as mode, melody, harmony, rhythm, and emotional expression based on music theory. Given that mode influences melody, and melody influences harmony, we design a sequential chat to link the mode agent, melody agent and harmony agent, ensuring they respond in reference to higher-level agents. The assessment criteria for each agent are detailed in \cref{tab:assess_agent}.

Next, we apply a multi-agent system for arrangement and mix. Since this process is challenging even for human musicians, we utilize prompt engineering techniques, including Role-play, Chain of Thought (CoT), and Few-Shot Prompting, instructing a group chat for this task. The agents receive prompts consisting of motion descriptions and melody in ABC notation, and collaborate on the final arrangement, with each agent's role outlined in \cref{tab:arrange_agent}.
%
%------------------------------------------------------------------------
\begin{table}[t]
  \centering
  \small
  \itshape
  \setlength{\abovecaptionskip}{1mm}
  \begin{tabular}{@{}p{1.2cm} p{6.3cm}@{}}
    \toprule
    Name & Assessment Criteria  \\
    \midrule
    Mode & 1.The mode is clear; 2.The tonality is stable. \\ \hline
    Melody & 3.Reasonable chord progression; 4.The chords are diverse, not monotonous; 5.Appropriate variation; 6.Appropriate repetition.  \\ \hline
    Harmony & 7.very harmonious; 8.rich, not monotonous; 9.The orchestration is reasonable; 10.the instruments collaborate well.\\ \hline
    Rhythm & 11.the rhythm is clear; 12.the beat is consistently the same. 13.The rhythmic pattern has appropriate variation; 14. The rhythmic pattern has appropriate repetition. \\ \hline
    Emotion & 15. The mode matches the emotion; 16. The chord progression matches the emotion; 17.the rhythm matches the emotion; 18. The choice of instruments matches the emotion; 19. The playing techniques fit the emotion (e.g., staccato, legato). \\
    \bottomrule
  \end{tabular}
  \caption{Agent names and their corresponding evaluation criteria.}
  \vspace{-2mm}
  \label{tab:assess_agent}
\end{table}
%------------------------------------------------------------------------
%
\begin{table}[t]
  \centering
  \small
  \itshape
  \setlength{\abovecaptionskip}{1mm}
  \begin{tabular}{@{}p{1.2cm} p{6.3cm}@{}}
    \toprule
    Name & Duties\\
    \midrule
    Analyze & 1.Summarize the development characteristics of the video; 2.Specify whether each measure should be forte or piano.  \\ \hline
    Arrange & 1.Specify which measures of which tracks should be softened; 2.Duplicate tracks to increase harmony. \\ \hline
    Instrument&Assign appropriate instrument.\\ \hline
    volume&Design the volume envelope.\\ \hline
    Mixing&1.Set the pan; 2.Set the reverb level.\\ \hline
    Reviewer&1.Review whether the arrangement and mix match the video in every aspect. 2.Examine the playing techniques and effectiveness of the instruments. 3.Develop the final scheme.\\
    \bottomrule
  \end{tabular}
  \caption{Arrangement and mix agent names and their duties.}
  \vspace{-6mm}
  \label{tab:arrange_agent}
\end{table}

In the group chat, the speaking order of each agent follows the actual sequence of music production, namely, analyzing the needs first, arranging and orchestrating next, then adjusting the volume to enhance dynamics, and finally mixing. Besides, the process of orchestration and dynamic adjustment needs to be guided by the reviewer agent.

\begin{figure}[ht]
    \centering
    \setlength{\abovecaptionskip}{1mm}
    \includegraphics[width=1\linewidth]{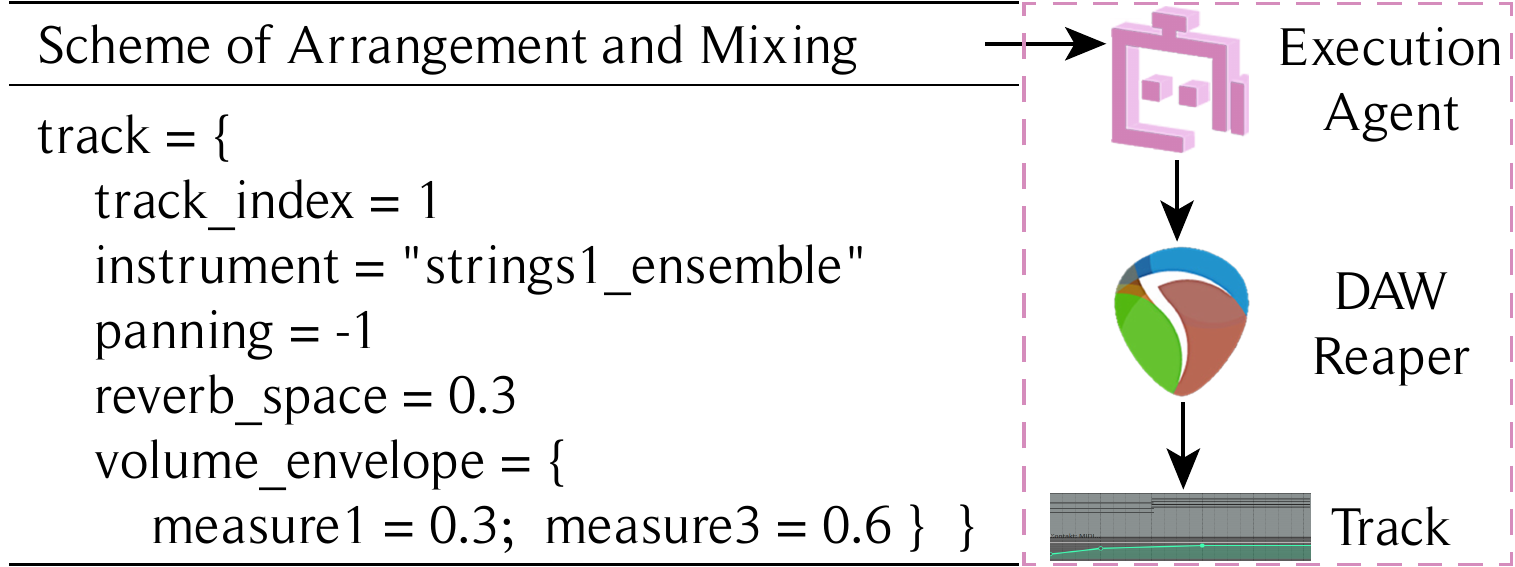}
    \caption{The output scheme from the multi-agent arrangement and mix system, together with subsequent operation.} % 整体大标题
    \vspace{-2mm}
    \label{fig:output-results}
\end{figure}

Since some tasks are difficult for the agent, CoT is used to break down steps and improve performance. For example, it is observed that the agent is somewhat sluggish in calculating, so the Analyze-Agent is instructed to first count harmonies per measure before further analyses. To prevent excessive instrumentation, the Instrument-Agent summarizes the original instrument types and makes selections accordingly. To guide agents' responses and ensure consistency, Few-Shot Prompting provides response templates, resulting in high-quality, well-structured outputs. The output scheme of arrangement and mix is shown in \cref{fig:output-results}.

Finally, an execution agent will convert this scheme into a form that Reaper---a lightweight, edit-friendly digital audio workstation---can understand, and thus guide Reaper to arrange and mix the melody of MIDI form by coding. Once completed, Reaper will output a piece of cinematic-grade music of exceptional quality, rich in musicality and perfectly aligned with the development of the film clips!
%------------------------------------------------------------------------
\section{MusicPro-7k Dataset}
\label{sec:dataset}

Given the lack of a suitable dataset to train FilmComposer, we construct a large-scale film music dataset MusicPro-7k featuring about 7,418 samples, each with film clip, high-quality music, visual description, music description, main melody, and rhythm spots. MusicPro-7k is distinguished by its high quality, high level of professional expertise, and multi-functionality. The structure and construction method of MusicPro-7k are illustrated in \cref{fig:dataset}, and more details can be found in supplementary materials.

\begin{figure}[t]
  \centering
  \includegraphics[width=0.48\textwidth]{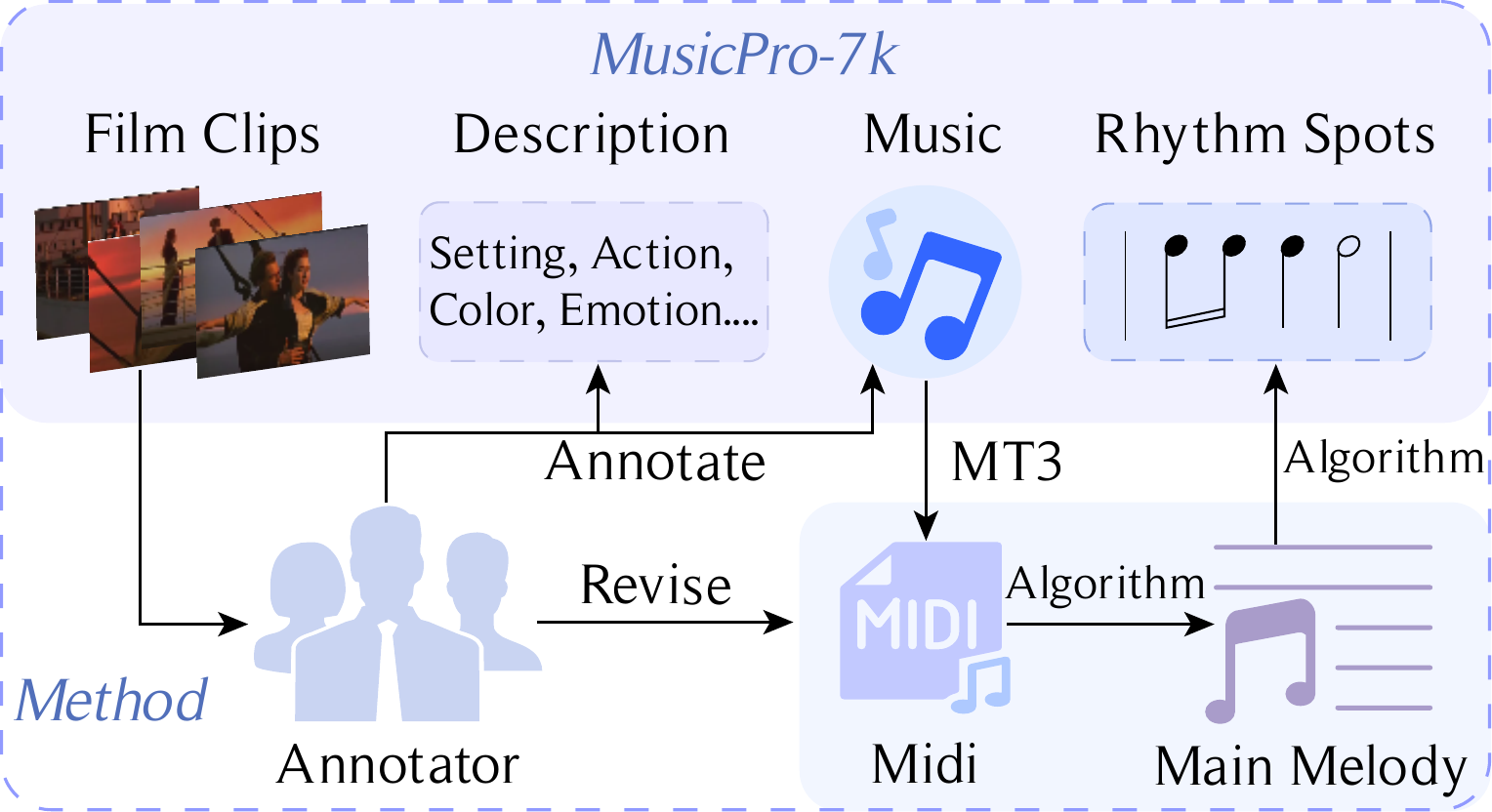}
   \setlength{\abovecaptionskip}{-2mm}
   \caption{The structure and construction method of MusicPro-7k, which consists of film clips, description, music and rhythm spots.}
   \label{fig:dataset}
   \vspace{-3mm}
\end{figure}

Instead of extracting music from videos and applying noise reduction as done in earlier approaches, we invited musicians to find music specifically for silent film clips. This approach avoids the poor audio quality and degradation caused by previous extraction methods.

\noindent \textbf{Film Clips.} 7K Film clips are derived from LSMDC \cite{rohrbach2017movie}, selecting video segments over 8 seconds long from numerous classic films. The source films cover all common genres, including drama, thriller, comedy, romance, and more.

\noindent \textbf{Music.} Musicians followed these steps: (1) For clips with pre-existing music, source their original audio; (2) If the original music could not be located or was under copyright protection, select a similar piece based on theoretical similarity and functional consistency as criteria; (3) For clips without pre-existing music, select suitable music from the music library, based on semantic, rhythmic, and progressive alignment with the clip, as well as all other relevant aspects.

\noindent \textbf{Description.} Musicians were also asked to annotate visual descriptions for each video based on predefined visual attributes, referring to audiovisual language theory and presented in supplementary materials. During the rhythm-controllable MusicGen fine-tuning, using the full visual description directly proved too broad in scope. To address this, we introduced music descriptions and removed components in stages, creating a smoother training process. The music descriptions, also annotated by the musicians, covered aspects such as genre, instrumentation, mood, and tempo.

\noindent \textbf{Rhythm Spots.} We first obtained preliminary main melodies and rhythm spots by transcribing the music and applying algorithms. Musicians then refined these using professional music production software, achieving high-quality, highly aligned results. This part of the dataset serves as key data for guiding large language models in rhythm alignment, marking a pioneering effort that we believe will significantly advance this field in the future.
%----------------------------------------------------------------------

\begin{table*}[ht]
  \centering
  \setlength{\abovecaptionskip}{1.5mm}
  \small 
  \tabcolsep=0.28cm
  \begin{tabular}{l|cccccccccc} 
  \midrule
   Method& KL$\downarrow$& FAD$\downarrow$& SR$\uparrow$& ImageBind$\uparrow$  &Diversity$\uparrow$ &Musicality$\uparrow$& Rhythm$\uparrow$& Dynamic$\downarrow$& Instru.$\downarrow$  \\ 
  \midrule
    GT & 0.000&0.000& 48K&0.328&0.451&14.40&4042&0.000&0.000\\   
    CMT\hfill\cite{di2021video} &1.554&0.644&-&0.104&0.361 &10.02&3153&0.801& 0.519\\    
    Video2Music\hfill\cite{kang2024video2music} &1.435&0.987&-&0.025&0.335&9.84&2001&1.018&0.535\\ M2UGen\hfill\cite{liu2023m}&1.569&0.306&32K&0.114&0.373&9.40&3392&1.070&0.510\\  VidMuse\hfill\cite{tian2024vidmuse}&2.035&0.376&32K&0.070&0.153&8.02&2467&0.892&0.527 \\
    MusicGen\hfill\cite{copet2024simple} &1.368&0.456&32K&0.108&0.133 &8.80&3050&1.147&0.546 \\ 
    \midrule
    RC-MusicGen & 1.320 & 0.219&32K&0.120&0.385&9.42&3646&0.820&0.506\\
    FilmComposer &\textbf{1.209}&\textbf{0.207}&\textbf{48K}&\textbf{0.131}&\textbf{0.444}&\textbf{10.78}&\textbf{3834}&\textbf{0.767} &\textbf{0.434}\\
  \bottomrule
  \end{tabular}
\caption{\textbf{Quantitative Results.} The metrics from left to right are: KL, FAD, Sampling Rate, ImageBind ranking scores, Diversity, Musicality, Rhythm Control, Dynamic Variation, Instrumentation. RC-MusicGen is the abbreviation for Rhythm-Controllable MusicGen.}
\vspace{-3mm}
\label{tab:objective}
\end{table*}

\section{Experiment}
\label{sec:ex}
\subsection{Experiment Setup}
We trained FilmComposer on MusicPro-7k, continuing from  MusicGen-Melody \cite{copet2024simple}. It converges around 150 epochs on dadam optimizer \cite{nazari2022dadam} with 2000 updates per epoch. The batch size is 4 and learning rate starts at e-1 and smaller after. Gradient clipping is applied with a maximum norm of 1.0. The Adam optimizer is configured with betas [0.9, 0.95], weight decay of 0.1, and epsilon set to 1e-8. A cosine learning rate schedule is employed, incorporating a warm-up phase of 4,000 steps, 0.0 minimum learning rate ratio and 1.0 cycle length. We also used cache that stored audio input to improve the training process. The training takes approximately six days using one NVIDIA A6000 GPU. During the evaluation phase, we established a test set that encompasses all thematic styles, ensuring the fairness of the evaluation process. 

\subsection{Experiment Evaluations}
\label{sec:metrics}

\noindent \textbf{For Music Quality}, we use the Fréchet Audio Distance (FAD) \cite{kilgour2018fr} to measure the distance between feature distributions of real and generated audio. We also use the sampling rate to intuitively reflect whether the audio quality has reached a usable standard.

\noindent \textbf{For Video-Music Correspondence}, we use ImageBind ranking scores\cite{girdhar2023imagebind} to assess relative rankings of audiovisual consistency, adhering to the official procedure of applying the softmax operation.

\noindent \textbf{For Diversity}, where the attraction of music lies, we propose a new metric \underline{\textit{Chroma-based Diversity}} to reflect. Chroma-based Diversity $D$ is calculated by averaging the pairwise differences between all distinct pairs:
\begin{equation} 
   D = 1 - \frac{2}{N(N-1)} \sum_{i=1}^{N} \sum_{j=i+1}^{N} Similarity(C_i, C_j), 
    \label{eq-diversity} 
\end{equation}
where $N$ is the total number of chroma feature matrices; $C_i$ is the $i$-th chroma feature matrix in the set; $Similarity(C_i, C_j)$ is the overall similarity over all segments between $C_i$ and $C_j$.

Previous work focuses on enhancing the quality and rhythm control of music generation, while they ignore the musicality and development of the music. As a more specialized form of music, the musicality and developmental aspects are particularly significant for film music.  

\noindent \textbf{For Musicality}, We propose a new metric named \underline{\textit{multi-agent music proficiency evaluation}} to measure the musicality of music generated. In detail, the music will be transformed to ABC notation and then be delivered to the multi-agent, being evaluated in terms of mode, melody, harmony, rhythm and emotional expression. We conduct user study to certify the accuracy of this metric.  

\noindent \textbf{For development}, we select three quantifiable aspects for evaluation---\underline{\textit{Rhythm control}}, \underline{\textit{Dynamic Variation}}, \underline{\textit{Instrumentation}}. The measurement of rhythm control involves using Madmom \cite{bock2016madmom} to extract beat points---acoustic events occurring in the audio, such as drum hits or the onset of notes---and then comparing them using cross-correlation. A higher value in the cross-correlation indicates better rhythm control. To elucidate, cross-correlation assesses the similarity between two signals by shifting one signal relative to the other and calculating the dot product at each shift position, thus identifying whether the two signals exhibit similar patterns or shapes at specific points. Then it can be calculated as: 
\begin{equation} 
    R_{xy}(k) = \sum_{t} x(t) y(t + k),
    \label{eq-rhythm} 
\end{equation}
where \( x(t) \) and \( y(t) \) represent two audio sequences; \( k \) is the lag applied to \( y(t) \) relative to \( x(t) \).

To measure Dynamic Variation, decibel changes are used to represent dynamics, where lower cosine similarity values indicate greater similarity in dynamic variation. For Instrumentation, the Musicnn \cite{pons2019musicnn} model with certain non-instrument tags removed is used to extract the distribution of instruments in the music, and where lower cosine similarity values indicate greater similarity in instrumentation.

\begin{figure}[t]
  \centering
  \setlength{\abovecaptionskip}{2mm}
  \includegraphics[width=1\linewidth]{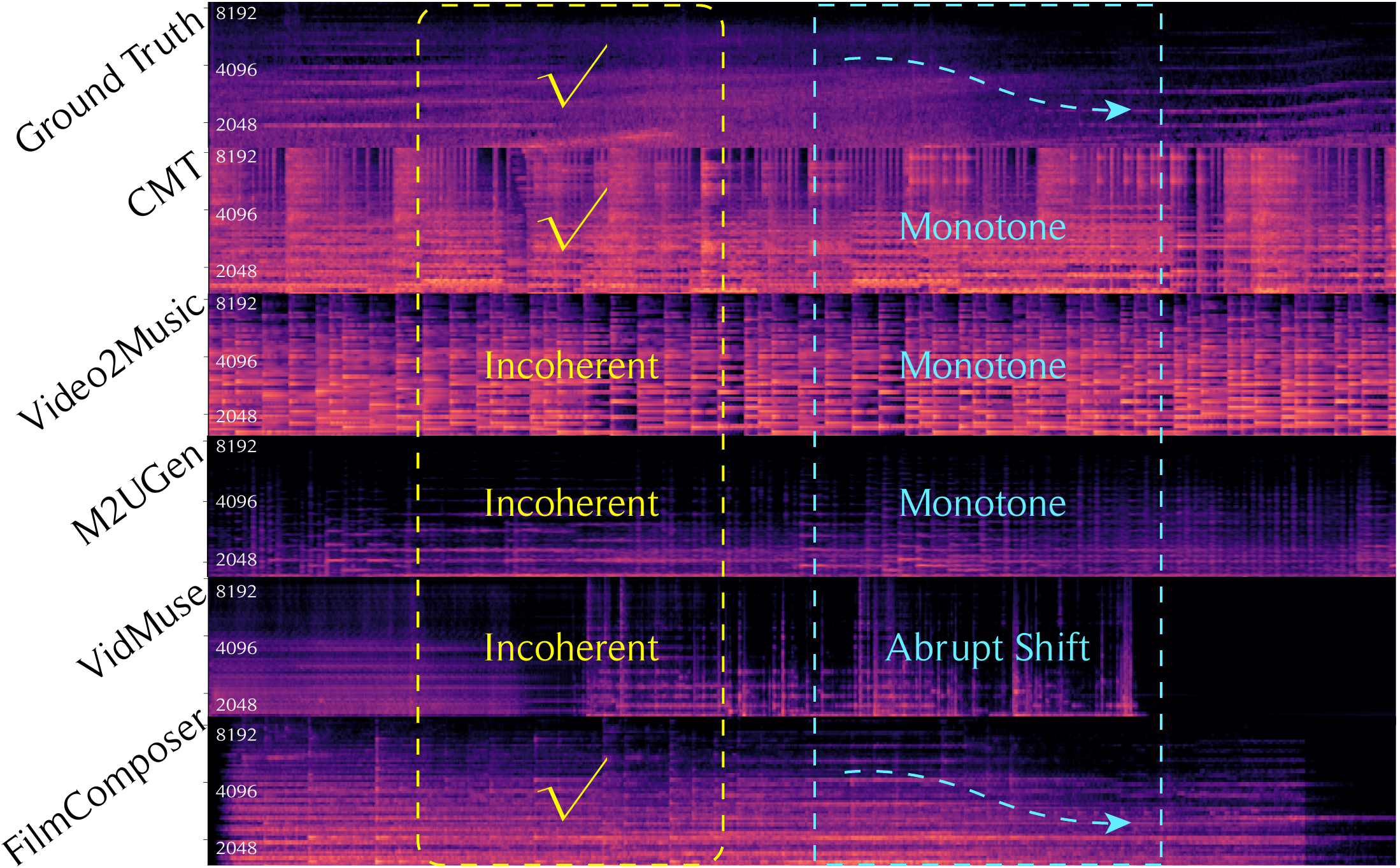}
   \caption{Qualitative Comparison results on spectrograms. The yellow box indicates where the music generated by Video2Music, M2UGen and VidMuse exhibits incoherence. The blue boxes show that CMT, Video2Music and M2UGen are generally monotone, while VidMuse presents abrupt shifts. In contrast, the ground truth and FilmComposer demonstrate clear layering.}
   \vspace{-4mm}
   \label{fig:qualitative}
\end{figure}

\subsection{Experiments Results}
% \begin{itemize}[leftmargin=1em] 

\noindent \textbf{Quantitative Analysis.}
As shown in \cref{tab:objective}, FilmComposer reaches state-of-the-art performance in every aspect.
%Some methods were not experimented with due to the lack of key model weights or functional code. 
Firstly, FilmComposer gets the lowest KL and rhythm-controllable MusicGen is second only to it, meaning that our method generates the most similar results to ground truth. The lowest FAD and the highest sampling rate indicate the high audio quality of final results generated by FilmComposer. The sampling rate has reached the cinematic standard after multi-agent arranging and mixing, which other models cannot implement. CMT and Video2Music generate symbolic music that has no sampling rate. Next, the higher ImageBind ranking scores than in other works proves that FilmComposer reaches the highest relevance between music and film clips. Rhythm-controllable MusicGen has already achieved a high diversity score, and with the addition of multi-agent arrangement, FilmComposer further improves it significantly, reaching a level close to that of the ground truth, indicating that our results exhibit excellent richness and diversity.

Besides these basic aspects, music generated by FilmComposer also reaches the greatest level of musicality and development. Since CMT and Video2Music generate symbolic music, their results are more accurate when transcribed into ABC notation, giving them an advantage in notation-based musicality evaluation. However, despite FilmComposer generating waveform music, it ultimately surpasses all other methods in this regard, demonstrating its strong musicality in music generated. Next, the advantages of musical development are reflected in three aspects---rhythm control, dynamic variation, and instrumentation. Rhythm control is one of the major focuses of our work, and the metrics show that FilmComposer surpasses CMT and VidMuse which also focus on rhythm control, demonstrating the precision and effectiveness of our approach in rhythm control. For dynamic variation similarity, we find that CMT is a strong competitor and rhythm-controllable MusicGen, which does not specifically address dynamics, performed worse than CMT. However, FilmComposer ultimately surpassed it, achieving the best dynamic variation. In terms of instrumentation similarity, FilmComposer far outperforms all other methods, proving the notable contribution of the multi-agent arrangement in instrumentation.

\noindent \textbf{Qualitative Analysis.}
Our qualitative analysis, presented in \cref{fig:qualitative}, highlights some limitations in other works. For CMT \cite{di2021video}, the music is generally monotone. Similarly, Video2Music \cite{kang2024video2music} produces monotonous outputs with excessive repetition. M2UGen's \cite{liu2023m} music cannot develop with video. VidMuse's \cite{kang2024video2music} music lacks substantive melody, with excessive noise and abnormal silences. Instead, our results are smooth and most similar to the ground truth in terms of rhythm and dynamics.

\subsection{Ablation Studies}
The rhythm-controllable MusicGen takes rhythm spots and description as conditions, so we conduct ablation studies to prove the effectiveness of these two modules. Additionally, we verify the crucial role of the multi-agent system in FilmComposer. The rhythm control modules, text control modules and multi-agent are removed in turn, and the evaluation results are shown in \cref{tab:ablation}.

\noindent \textbf{Text control.} 
When the text control module is added, FAD metrics improve largely, which indicates increasing audio quality. It's noteworthy that the text control module also helps improve the rhythm control, as visual description can help the model decide the proper rhythm patterns. When the text control module is working, ImageBind gets a very high ranking score, proving its notable contribution to enhancing the correspondence between music and film clips.

\noindent \textbf{Rhythm control.} 
After adding rhythm control module, the rhythm control metrics has been greatly improved, proving rhythm condition's significance. The rhythm control module also helps increase the audio quality and correspondence since the FAD metrics and ImageBind ranking rate improve.

\noindent \textbf{Multi-agent.} 
The inclusion of the agent led to improvements in FAD, Rhythm control, ImageBind ranking score, proving its effectiveness in enhancing audio quality, rhythm control and visual-audio relevance.

\begin{table}[t]
  \centering
    \setlength{\abovecaptionskip}{1.5mm}
  \small 
  \tabcolsep=0.25cm
  \begin{tabular}{c|c|c|ccc}
    \toprule
    Text & Rhythm & Agent & FAD$\downarrow$ &Rhythm$\uparrow$& ImageBind$\uparrow$ \\
    \midrule
    $\checkmark$ &$\times$ &$\times$ & 0.246 & 2978&0.254\\
    $\times$ &$\checkmark$ & $\times$ & 0.319 & 2836&0.163\\
    $\checkmark$& $\checkmark$ &$\times$ & 0.219 &3646&0.265\\
    $\checkmark$& $\checkmark$ &$\checkmark$ & \textbf{0.207} &\textbf{3834}&\textbf{0.318}\\
    
    \bottomrule
  \end{tabular}
  \caption{\textbf{Results of ablation study.} FAD validates improvement in audio quality, Rhythm control metrics assess the effectiveness of rhythm control, and ImageBind ranking scores verify the enhancement of visual-audio relevance.}
  \vspace{-4.5mm}
  \label{tab:ablation}
\end{table}
%----------------------------------------------------------------------
\subsection{User Study}
The best way to measure the professionalism of music production remains user study. We invited a total of 51 people, including 18 experts with professional backgrounds in film music production and over five years of professional experience, to evaluate music generated by FilmComposer, CMT \cite{di2021video} and Video2Music \cite{kang2024video2music} in several aspects and give their preference separately. The aspects are: (1) Musicality: the innate ability to perceive, understand, and express musical elements such as mode, melody, harmony and rhythm; (2) Content Correspondence: The match between music and video in terms of emotion and semantics; (3) Rhythm correspondence: the alignment of rhythmic patterns with video; (4) Dynamic correspondence: the alignment of varying levels of loudness and intensity with video; (5) Instrumentation correspondence: the alignment of instrument usage with video. (6) Overall: the general assessment of music.

\noindent \textbf{Results.} 
\cref{tab:user} shows the results. More than 60\% of both experts and non-experts preferred the music generated by FilmComposer for each item. This demonstrates FilmComposer's significant advantages in musicality, multiple aspects of correspondence, and overall quality. The user study yield results consistent with previous experiments, validating the reliability of our newly proposed metrics in \cref{sec:metrics}.

\begin{table}[t]
  \centering
  \small 
  \setlength{\abovecaptionskip}{1mm}
  \tabcolsep=0.6cm
  \begin{tabular}{l|ccc}
    \toprule
    Metrics & Experts & Non-Experts \\
    \midrule
    Musicality &64.4\%&78.8\%\\
    C.Content &82.2\%&90.9\% \\
    C.Rhythm &73.3\%&86.7\%\\
    C.Dynamic &75.6\%&-\\
    C.Instru &78.9\%&-\\
    Overall &81.1\%&90.3\%\\
    \bottomrule
  \end{tabular}
  \caption{\textbf{Results of user study.} The data indicate the percentage of participants favoring FilmComposer. Non-Experts evaluate intuitive audiovisual alignment, without needing to consider more specialized musical knowledge. C. represents for Correspondence.}
  \vspace{-4.5mm}
  \label{tab:user}
\end{table}
%----------------------------------------------------------------------

\section{Application}
\label{sec:app}
Since our framework imitates musicians, it can integrate into the actual music production process seamlessly and easily. So we design a system for convenient human-computer interaction, allowing users to edit and optimize almost every step to create music tailored to their needs. More details can be found in supplementary materials. \textbf{(1) Rhythm spotting:} Users can freely and conveniently drag, delete, or add rhythm spots to create their desired pattern, since rhythm is represented by a series of spots. \textbf{(2) Content descriptions:} The visual and motion descriptions are both in text forms, easy for users to modify. \textbf{(3) Musical notation:} After the music is transcribed into ABC notation, users can directly edit it to gain their desired melody. \textbf{(4) Arrangement and mix scheme:} Users can modify it before it is executed by the Execution Agent.

Non-experts can also use our system to easily and conveniently produce various types of music, such as background music for personal projects or social media content. The system can also serve educational purposes, helping beginners quickly understand the music production process and learn techniques through hands-on experimentation.

\section{Conclusion}
\label{sec:conclusion}
In this work, we propose the FilmComposer framework and the first film music dataset MusicPro-7k, implementing the film music production task and resulting in state-of-the-art performance. Our LLM-driven innovative approach has been proven highly effective across multiple metrics, which imitates musicians to conduct music production and combines symbolic and waveform music generation. Finally, the music produced by our method meets a cinematic standard suitable for direct use, exhibiting high quality, strong musicality, development, correspondence with film clips, and diversity, effectively addressing the challenges of high professionalism in film music production. FilmComposer can seamlessly integrate into the actual music production process, allowing user intervention in various aspects.
%such as rhythm spotting, content description, transcription results, arrangement and mix. 
This provides strong interactivity and a high degree of creative freedom. For discussions on model analysis, extensions, and limitations, please refer to the supplementary materials.

%\noindent \textbf{Acknowledgments.} This work is supported by the National Natural Science Foundation of China (No.62402306) and Natural Science Foundation of Shanghai (No.24ZR1422400).

%, and designs a multi-agent collaborative system integrated with a digital audio workstation.
\section*{Acknowledgments}
\label{sec:acknowledgments}
This work is supported by the National Natural Science Foundation of China (No.62402306) and Natural Science Foundation of Shanghai (No.24ZR1422400).
{
    \small
    \bibliographystyle{ieeenat_fullname}
    \bibliography{main}

\begin{thebibliography}{45}
\providecommand{\natexlab}[1]{#1}
\providecommand{\url}[1]{\texttt{#1}}
\expandafter\ifx\csname urlstyle\endcsname\relax
  \providecommand{\doi}[1]{doi: #1}\else
  \providecommand{\doi}{doi: \begingroup \urlstyle{rm}\Url}\fi

\bibitem[Achiam et~al.(2023)Achiam, Adler, Agarwal, Ahmad, Akkaya, Aleman, Almeida, Altenschmidt, Altman, Anadkat, et~al.]{achiam2023gpt}
Josh Achiam, Steven Adler, Sandhini Agarwal, Lama Ahmad, Ilge Akkaya, Florencia~Leoni Aleman, Diogo Almeida, Janko Altenschmidt, Sam Altman, Shyamal Anadkat, et~al.
\newblock Gpt-4 technical report.
\newblock \emph{arXiv preprint arXiv:2303.08774}, 2023.

\bibitem[Agostinelli et~al.(2023)Agostinelli, Denk, Borsos, Engel, Verzetti, Caillon, Huang, Jansen, Roberts, Tagliasacchi, et~al.]{agostinelli2023musiclm}
Andrea Agostinelli, Timo~I Denk, Zal{\'a}n Borsos, Jesse Engel, Mauro Verzetti, Antoine Caillon, Qingqing Huang, Aren Jansen, Adam Roberts, Marco Tagliasacchi, et~al.
\newblock Musiclm: Generating music from text.
\newblock \emph{arXiv preprint arXiv:2301.11325}, 2023.

\bibitem[B{\"o}ck et~al.(2016)B{\"o}ck, Korzeniowski, Schl{\"u}ter, Krebs, and Widmer]{bock2016madmom}
Sebastian B{\"o}ck, Filip Korzeniowski, Jan Schl{\"u}ter, Florian Krebs, and Gerhard Widmer.
\newblock Madmom: A new python audio and music signal processing library.
\newblock In \emph{Proceedings of the 24th ACM international conference on Multimedia}, pages 1174--1178, 2016.

\bibitem[Bradski(2000)]{bradski2000opencv}
G Bradski.
\newblock The opencv library.
\newblock \emph{Dr. Dobb’s Journal of Software Tools}, 2000.

\bibitem[Breakthrough(2023)]{SceneDetect}
Breakthrough.
\newblock Pyscenedetect: Video scene cut detection tool, 2023.
\newblock Version 0.6, Available at \url{https://github.com/Breakthrough/PySceneDetect}.

\bibitem[Cheng et~al.(2024)Cheng, Lu, Li, Zai, Yin, Cheng, Yan, and Liang]{cheng2024autostudio}
Junhao Cheng, Xi Lu, Hanhui Li, Khun~Loun Zai, Baiqiao Yin, Yuhao Cheng, Yiqiang Yan, and Xiaodan Liang.
\newblock Autostudio: Crafting consistent subjects in multi-turn interactive image generation.
\newblock \emph{arXiv preprint arXiv:2406.01388}, 2024.

\bibitem[Copet et~al.(2024)Copet, Kreuk, Gat, Remez, Kant, Synnaeve, Adi, and D{\'e}fossez]{copet2024simple}
Jade Copet, Felix Kreuk, Itai Gat, Tal Remez, David Kant, Gabriel Synnaeve, Yossi Adi, and Alexandre D{\'e}fossez.
\newblock Simple and controllable music generation.
\newblock \emph{Advances in Neural Information Processing Systems}, 36, 2024.

\bibitem[Deng et~al.(2024)Deng, Yang, Yuan, Huang, Wang, Liu, Tian, Pan, Zhang, Lin, et~al.]{deng2024composerx}
Qixin Deng, Qikai Yang, Ruibin Yuan, Yipeng Huang, Yi Wang, Xubo Liu, Zeyue Tian, Jiahao Pan, Ge Zhang, Hanfeng Lin, et~al.
\newblock Composerx: Multi-agent symbolic music composition with llms.
\newblock \emph{arXiv preprint arXiv:2404.18081}, 2024.

\bibitem[Di et~al.(2021)Di, Jiang, Liu, Wang, Zhu, He, Liu, and Yan]{di2021video}
Shangzhe Di, Zeren Jiang, Si Liu, Zhaokai Wang, Leyan Zhu, Zexin He, Hongming Liu, and Shuicheng Yan.
\newblock Video background music generation with controllable music transformer.
\newblock In \emph{Proceedings of the 29th ACM International Conference on Multimedia}, pages 2037--2045, 2021.

\bibitem[Dong et~al.(2024)Dong, Liu, Chen, Polak, and Zhang]{dong2024musechat}
Zhikang Dong, Xiulong Liu, Bin Chen, Pawel Polak, and Peng Zhang.
\newblock Musechat: A conversational music recommendation system for videos.
\newblock In \emph{Proceedings of the IEEE/CVF Conference on Computer Vision and Pattern Recognition}, pages 12775--12785, 2024.

\bibitem[Evans et~al.(2024)Evans, Carr, Taylor, Hawley, and Pons]{evans2024fast}
Zach Evans, CJ Carr, Josiah Taylor, Scott~H Hawley, and Jordi Pons.
\newblock Fast timing-conditioned latent audio diffusion.
\newblock \emph{arXiv preprint arXiv:2402.04825}, 2024.

\bibitem[Gan et~al.(2020)Gan, Huang, Chen, Tenenbaum, and Torralba]{gan2020foley}
Chuang Gan, Deng Huang, Peihao Chen, Joshua~B Tenenbaum, and Antonio Torralba.
\newblock Foley music: Learning to generate music from videos.
\newblock In \emph{Computer Vision--ECCV 2020: 16th European Conference, Glasgow, UK, August 23--28, 2020, Proceedings, Part XI 16}, pages 758--775. Springer, 2020.

\bibitem[Gardner et~al.(2021)Gardner, Simon, Manilow, Hawthorne, and Engel]{gardner2021mt3}
Josh Gardner, Ian Simon, Ethan Manilow, Curtis Hawthorne, and Jesse Engel.
\newblock Mt3: Multi-task multitrack music transcription.
\newblock \emph{arXiv preprint arXiv:2111.03017}, 2021.

\bibitem[Girdhar et~al.(2023)Girdhar, El-Nouby, Liu, Singh, Alwala, Joulin, and Misra]{girdhar2023imagebind}
Rohit Girdhar, Alaaeldin El-Nouby, Zhuang Liu, Mannat Singh, Kalyan~Vasudev Alwala, Armand Joulin, and Ishan Misra.
\newblock Imagebind: One embedding space to bind them all.
\newblock In \emph{CVPR}, 2023.

\bibitem[Guzhov et~al.(2022)Guzhov, Raue, Hees, and Dengel]{guzhov2022audioclip}
Andrey Guzhov, Federico Raue, J{\"o}rn Hees, and Andreas Dengel.
\newblock Audioclip: Extending clip to image, text and audio.
\newblock In \emph{ICASSP 2022-2022 IEEE International Conference on Acoustics, Speech and Signal Processing (ICASSP)}, pages 976--980. IEEE, 2022.

\bibitem[Han et~al.(2022)Han, Kong, Han, and Wang]{han2022survey}
Donghong Han, Yanru Kong, Jiayi Han, and Guoren Wang.
\newblock A survey of music emotion recognition.
\newblock \emph{Frontiers of Computer Science}, 16\penalty0 (6):\penalty0 166335, 2022.

\bibitem[Hevner(1935)]{hevner1935affective}
Kate Hevner.
\newblock The affective character of the major and minor modes in music.
\newblock \emph{The American Journal of Psychology}, 47\penalty0 (1):\penalty0 103--118, 1935.

\bibitem[Huang et~al.(2023)Huang, Park, Wang, Denk, Ly, Chen, Zhang, Zhang, Yu, Frank, et~al.]{huang2023noise2music}
Qingqing Huang, Daniel~S Park, Tao Wang, Timo~I Denk, Andy Ly, Nanxin Chen, Zhengdong Zhang, Zhishuai Zhang, Jiahui Yu, Christian Frank, et~al.
\newblock Noise2music: Text-conditioned music generation with diffusion models.
\newblock \emph{arXiv preprint arXiv:2302.03917}, 2023.

\bibitem[Kang et~al.(2024)Kang, Poria, and Herremans]{kang2024video2music}
Jaeyong Kang, Soujanya Poria, and Dorien Herremans.
\newblock Video2music: Suitable music generation from videos using an affective multimodal transformer model.
\newblock \emph{Expert Systems with Applications}, 249:\penalty0 123640, 2024.

\bibitem[Kilgour et~al.(2018)Kilgour, Zuluaga, Roblek, and Sharifi]{kilgour2018fr}
Kevin Kilgour, Mauricio Zuluaga, Dominik Roblek, and Matthew Sharifi.
\newblock Fr$\backslash$'echet audio distance: A metric for evaluating music enhancement algorithms.
\newblock \emph{arXiv preprint arXiv:1812.08466}, 2018.

\bibitem[Lan et~al.(2024)Lan, Hsiao, Cheng, and Yang]{lan2024musicongen}
Yun-Han Lan, Wen-Yi Hsiao, Hao-Chung Cheng, and Yi-Hsuan Yang.
\newblock Musicongen: Rhythm and chord control for transformer-based text-to-music generation.
\newblock \emph{arXiv preprint arXiv:2407.15060}, 2024.

\bibitem[Lee et~al.(2022)Lee, Kim, Kang, Ki, Hwang, Han, Kim, et~al.]{lee2022commu}
Hyun Lee, Taehyun Kim, Hyolim Kang, Minjoo Ki, Hyeonchan Hwang, Sharang Han, Seon~Joo Kim, et~al.
\newblock Commu: Dataset for combinatorial music generation.
\newblock \emph{Advances in Neural Information Processing Systems}, 35:\penalty0 39103--39114, 2022.

\bibitem[Li et~al.(2021)Li, Yang, Ross, and Kanazawa]{li2021ai}
Ruilong Li, Shan Yang, David~A Ross, and Angjoo Kanazawa.
\newblock Ai choreographer: Music conditioned 3d dance generation with aist++.
\newblock In \emph{Proceedings of the IEEE/CVF International Conference on Computer Vision}, pages 13401--13412, 2021.

\bibitem[Li et~al.(2024)Li, Qin, Zheng, Jin, and Liu]{li2024diff}
Sizhe Li, Yiming Qin, Minghang Zheng, Xin Jin, and Yang Liu.
\newblock Diff-bgm: A diffusion model for video background music generation.
\newblock In \emph{Proceedings of the IEEE/CVF Conference on Computer Vision and Pattern Recognition}, pages 27348--27357, 2024.

\bibitem[Liu et~al.(2024)Liu, Yuan, Liu, Mei, Kong, Tian, Wang, Wang, Wang, and Plumbley]{liu2024audioldm}
Haohe Liu, Yi Yuan, Xubo Liu, Xinhao Mei, Qiuqiang Kong, Qiao Tian, Yuping Wang, Wenwu Wang, Yuxuan Wang, and Mark~D Plumbley.
\newblock Audioldm 2: Learning holistic audio generation with self-supervised pretraining.
\newblock \emph{IEEE/ACM Transactions on Audio, Speech, and Language Processing}, 2024.

\bibitem[Liu et~al.(2023)Liu, Hussain, Sun, and Shan]{liu2023m}
Shansong Liu, Atin~Sakkeer Hussain, Chenshuo Sun, and Ying Shan.
\newblock M$^2$ ugen: Multi-modal music understanding and generation with the power of large language models.
\newblock \emph{arXiv preprint arXiv:2311.11255}, 2023.

\bibitem[McKee et~al.(2023)McKee, Salamon, Sivic, and Russell]{mckee2023language}
Daniel McKee, Justin Salamon, Josef Sivic, and Bryan Russell.
\newblock Language-guided music recommendation for video via prompt analogies.
\newblock In \emph{Proceedings of the IEEE/CVF Conference on Computer Vision and Pattern Recognition}, pages 14784--14793, 2023.

\bibitem[Nazari et~al.(2022)Nazari, Tarzanagh, and Michailidis]{nazari2022dadam}
Parvin Nazari, Davoud~Ataee Tarzanagh, and George Michailidis.
\newblock Dadam: A consensus-based distributed adaptive gradient method for online optimization.
\newblock \emph{IEEE Transactions on Signal Processing}, 70:\penalty0 6065--6079, 2022.

\bibitem[Pons and Serra(2019)]{pons2019musicnn}
Jordi Pons and Xavier Serra.
\newblock musicnn: Pre-trained convolutional neural networks for music audio tagging.
\newblock \emph{arXiv preprint arXiv:1909.06654}, 2019.

\bibitem[Pr{\'e}tet et~al.(2021)Pr{\'e}tet, Richard, and Peeters]{pretet2021cross}
Laure Pr{\'e}tet, Gael Richard, and Geoffroy Peeters.
\newblock Cross-modal music-video recommendation: A study of design choices.
\newblock In \emph{2021 International Joint Conference on Neural Networks (IJCNN)}, pages 1--9. IEEE, 2021.

\bibitem[Raffel et~al.(2020)Raffel, Shazeer, Roberts, Lee, Narang, Matena, Zhou, Li, and Liu]{raffel2020exploring}
Colin Raffel, Noam Shazeer, Adam Roberts, Katherine Lee, Sharan Narang, Michael Matena, Yanqi Zhou, Wei Li, and Peter~J Liu.
\newblock Exploring the limits of transfer learning with a unified text-to-text transformer.
\newblock \emph{Journal of machine learning research}, 21\penalty0 (140):\penalty0 1--67, 2020.

\bibitem[Rohrbach et~al.(2017)Rohrbach, Torabi, Rohrbach, Tandon, Pal, Larochelle, Courville, and Schiele]{rohrbach2017movie}
Anna Rohrbach, Atousa Torabi, Marcus Rohrbach, Niket Tandon, Christopher Pal, Hugo Larochelle, Aaron Courville, and Bernt Schiele.
\newblock Movie description.
\newblock \emph{International Journal of Computer Vision}, 123:\penalty0 94--120, 2017.

\bibitem[Russell(1980)]{russell1980circumplex}
James~A Russell.
\newblock A circumplex model of affect.
\newblock \emph{Journal of personality and social psychology}, 39\penalty0 (6):\penalty0 1161, 1980.

\bibitem[Schneider et~al.(2024)Schneider, Kamal, Jin, and Sch{\"o}lkopf]{schneider2024mousai}
Flavio Schneider, Ojasv Kamal, Zhijing Jin, and Bernhard Sch{\"o}lkopf.
\newblock Mo{\^u}sai: Efficient text-to-music diffusion models.
\newblock In \emph{Proceedings of the 62nd Annual Meeting of the Association for Computational Linguistics (Volume 1: Long Papers)}, pages 8050--8068, 2024.

\bibitem[Sur{\'\i}s et~al.(2022)Sur{\'\i}s, Vondrick, Russell, and Salamon]{suris2022s}
D{\'\i}dac Sur{\'\i}s, Carl Vondrick, Bryan Russell, and Justin Salamon.
\newblock It's time for artistic correspondence in music and video.
\newblock In \emph{Proceedings of the IEEE/CVF Conference on Computer Vision and Pattern Recognition}, pages 10564--10574, 2022.

\bibitem[Tian et~al.(2024)Tian, Liu, Yuan, Pan, Huang, Liu, Tan, Chen, Xue, and Guo]{tian2024vidmuse}
Zeyue Tian, Zhaoyang Liu, Ruibin Yuan, Jiahao Pan, Xiaoqiang Huang, Qifeng Liu, Xu Tan, Qifeng Chen, Wei Xue, and Yike Guo.
\newblock Vidmuse: A simple video-to-music generation framework with long-short-term modeling.
\newblock \emph{arXiv preprint arXiv:2406.04321}, 2024.

\bibitem[Wu et~al.(2022)Wu, Seetharaman, Kumar, and Bello]{wu2022wav2clip}
Ho-Hsiang Wu, Prem Seetharaman, Kundan Kumar, and Juan~Pablo Bello.
\newblock Wav2clip: Learning robust audio representations from clip.
\newblock In \emph{ICASSP 2022-2022 IEEE International Conference on Acoustics, Speech and Signal Processing (ICASSP)}, pages 4563--4567. IEEE, 2022.

\bibitem[Wu et~al.(2023)Wu, Bansal, Zhang, Wu, Zhang, Zhu, Li, Jiang, Zhang, and Wang]{wu2023autogen}
Qingyun Wu, Gagan Bansal, Jieyu Zhang, Yiran Wu, Shaokun Zhang, Erkang Zhu, Beibin Li, Li Jiang, Xiaoyun Zhang, and Chi Wang.
\newblock Autogen: Enabling next-gen llm applications via multi-agent conversation framework.
\newblock \emph{arXiv preprint arXiv:2308.08155}, 2023.

\bibitem[xlvector(2016)]{midi2abc}
xlvector.
\newblock abcmidi.
\newblock Online, 2016.
\newblock \url{https://github.com/xlvector/abcmidi}.

\bibitem[Yi et~al.(2021)Yi, Zhu, Xie, and Chen]{yi2021cross}
Jing Yi, Yaochen Zhu, Jiayi Xie, and Zhenzhong Chen.
\newblock Cross-modal variational auto-encoder for content-based micro-video background music recommendation.
\newblock \emph{IEEE Transactions on Multimedia}, 25:\penalty0 515--528, 2021.

\bibitem[Yu et~al.(2023{\natexlab{a}})Yu, Song, Lu, He, Tan, Ye, Zhang, and Bian]{yu2023musicagent}
Dingyao Yu, Kaitao Song, Peiling Lu, Tianyu He, Xu Tan, Wei Ye, Shikun Zhang, and Jiang Bian.
\newblock Musicagent: An ai agent for music understanding and generation with large language models.
\newblock \emph{arXiv preprint arXiv:2310.11954}, 2023{\natexlab{a}}.

\bibitem[Yu et~al.(2023{\natexlab{b}})Yu, Wang, Chen, Sun, and Qiao]{yu2023long}
Jiashuo Yu, Yaohui Wang, Xinyuan Chen, Xiao Sun, and Yu Qiao.
\newblock Long-term rhythmic video soundtracker.
\newblock In \emph{International Conference on Machine Learning}, pages 40339--40353. PMLR, 2023{\natexlab{b}}.

\bibitem[Zeng et~al.(2018)Zeng, Yu, and Oyama]{zeng2018audio}
Donghuo Zeng, Yi Yu, and Keizo Oyama.
\newblock Audio-visual embedding for cross-modal music video retrieval through supervised deep cca.
\newblock In \emph{2018 IEEE International Symposium on Multimedia (ISM)}, pages 143--150. IEEE, 2018.

\bibitem[Zhu et~al.(2022)Zhu, Olszewski, Wu, Achlioptas, Chai, Yan, and Tulyakov]{zhu2022quantized}
Ye Zhu, Kyle Olszewski, Yu Wu, Panos Achlioptas, Menglei Chai, Yan Yan, and Sergey Tulyakov.
\newblock Quantized gan for complex music generation from dance videos.
\newblock In \emph{European Conference on Computer Vision}, pages 182--199. Springer, 2022.

\bibitem[Zhuo et~al.(2023)Zhuo, Wang, Wang, Liao, Bao, Peng, Han, Zhang, Fang, and Liu]{zhuo2023video}
Le Zhuo, Zhaokai Wang, Baisen Wang, Yue Liao, Chenxi Bao, Stanley Peng, Songhao Han, Aixi Zhang, Fei Fang, and Si Liu.
\newblock Video background music generation: Dataset, method and evaluation.
\newblock In \emph{Proceedings of the IEEE/CVF International Conference on Computer Vision}, pages 15637--15647, 2023.

\end{thebibliography}
}

% WARNING: do not forget to delete the supplementary pages from your submission 
\clearpage
\maketitlesupplementary

\renewcommand{\thesection}{\Alph{section}}
\setcounter{section}{0}
\renewcommand{\thesubsection}{\thesection.\arabic{subsection}}
\section{MusicPro-7k Details}
\begin{figure}[ht]
    \centering
    \vspace{-3mm}
    \includegraphics[width=1\linewidth]{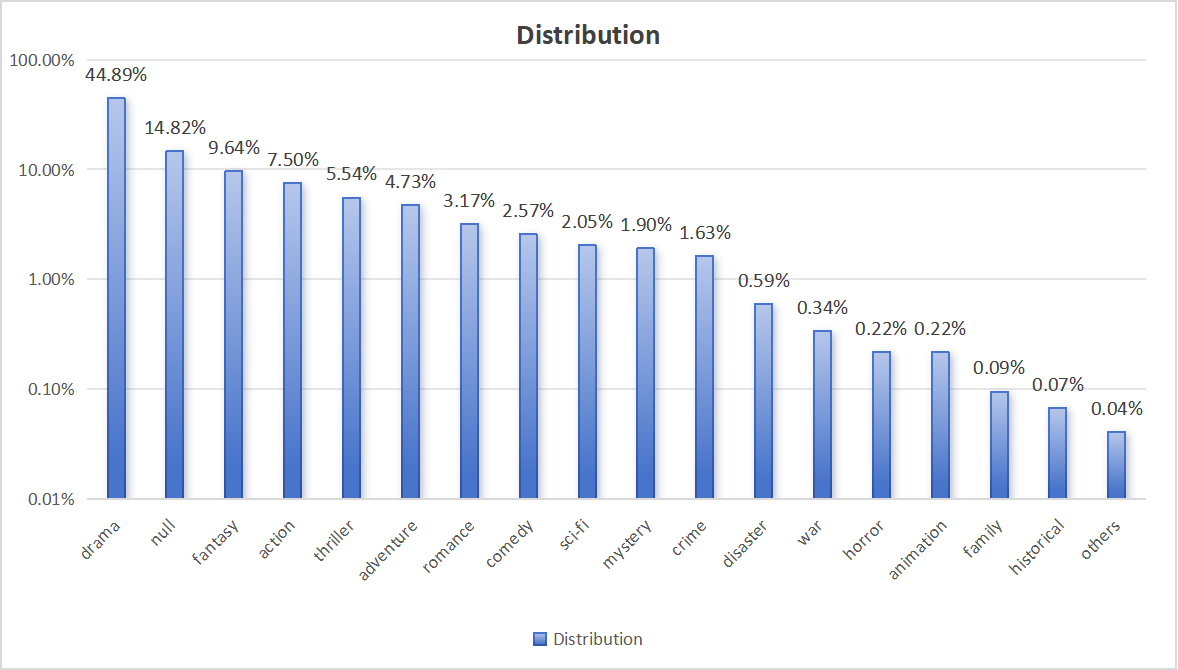}
    \captionof{figure}{The distribution of MusicPro-7k. ``Null'' represents film clips without a clearly defined theme, while ``others'' includes themes such as biography, sport, and documentary. The vertical axis in the figure uses a logarithmic scale.}
    \label{fig:distribution}
    \vspace{-2mm}
\end{figure}

Our dataset MusicPro-7k covers films of all common genres, and the distribution is shown in \cref{fig:distribution}. Since most film genres include drama and many clips prominently feature dramatic elements, the proportion of the drama theme in the MusicPro-7k is relatively high.

\vspace{5pt}

Musicians need to annotate visual descriptions, and to ensure the effectiveness of the annotations, predefined visual attributes need to be set to cover all film clips. The complete predefined visual attributes are present in \cref{tab:complete_description}.

\begin{table}[ht]
  \centering
  \small
  \itshape
 \setlength{\abovecaptionskip}{1.5mm}
  \begin{tabular}{@{}p{1cm} p{6.5cm}@{}}
    \toprule
    category & labels\\
    \midrule
    Setting & road, home, forest or jungle or canyon, corridor, car, sea, Kitchen and dining area, ruins, hospital, garden, space ship or outer space, hotel, workplace, bar, stairs, airport, city, ship, castle, prison, lab, beach, train station, desert, rooftop, bus, police station, elevator, stage, tunnel, alley, train, factory, square, school, playground, tent, theater, cave , church, bridge, pier, military base, night view, balcony, store, market, court, ballroom, under water, parking lot, swimming pool, rural, casino, grassland, library or bookstore, cemetery, farm, subway station, cliff, coffee shop, rain, street, null\\
    \midrule
    Brightness &  mild, bright, dull, somber, dark, glaring, contrasting\\
    \midrule
    Color Hue & Blue, Green, Red, Pink, Yellow, Orange, Purple, Black, White, Brown, Gray \\
    \midrule
    Behavior &  shoot gun, run, call, do intimacy, kiss, fight, drive car, read, drink, smoke, climb, fall down, ride horse, eat, applaud and cheer, dance, cry, hug, write, drive plane, work, sleep, laugh, quarrel, ride motorcycle, kill or attack, pursue and arrest, play instrument, swim, fire, faint, play ball games, take shower, play games, boating, cook, sing, ride bicycle, paint, do housework, pray, sit, talk, battle, speech, goodbye, gaze, escape, open door, null \\
    \midrule
    View Scale & long shot, full shot, medium shot, close-up shot, extreme close-up shot \\
    \midrule
    Emotion & Dignified, Sad, Dreamy, Calm, Graceful, Joyous, Exciting, Vigorous, Nervous, Angry, Fear, Humorous, null \\
    \midrule
    Theme & drama, fantasy, action, thriller, adventure, romance, comedy, sci-fi, mystery, crime, disaster, war, horror, animation, family, historical, biography, sport, documentary, null \\
    \bottomrule
  \end{tabular}
  \caption{The categories of visual descriptions and their corresponding labels.}
  \vspace{-5.5mm}
  \label{tab:complete_description}
\end{table}

\section{Experiment Details}
\subsection{Intermediate Results}
\begin{figure}[H]
    \centering
    \setlength{\abovecaptionskip}{-1mm}
    \begin{minipage}[t]{0.45\linewidth} % 左边两张图的容器
        \centering
        \subcaptionbox{ABC Notation}[1\linewidth]{ % 第二张图的标题
            \includegraphics[width=\linewidth]{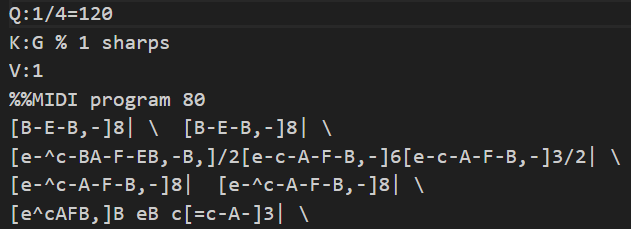} % 替换为第二张图片路径
        }
        \vspace{0.5em} % 两张图片之间的间距
        \subcaptionbox{Output of Evaluation Agent}[1\linewidth]{ % 第一张图的标题
            \includegraphics[width=\linewidth]{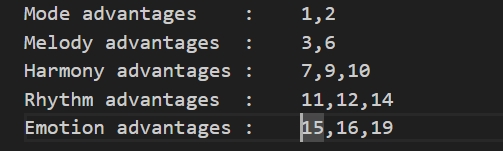} % 替换为第一张图片路径
        }
    \end{minipage}
    \hfill
    \raisebox{-0.54\height}{ % 手动调整右侧图片的垂直位置
        \begin{minipage}[t]{0.5\linewidth} % 右边单张图的容器
            \centering
            \subcaptionbox{Arrangement and Mix Scheme}[1\linewidth]{ % 第三张图的标题
                \includegraphics[width=\linewidth]{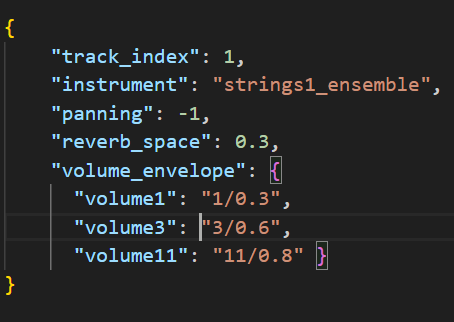} % 替换为第三张图片路径
            }
        \end{minipage}
    }
    \caption{The output results are respectively from the transcription, multi-agent assessment system, and multi-agent arrangement and mixing system.} % 整体大标题
    \vspace{-2mm}
    \label{fig:stage_results}
\end{figure}

In multi-agent assessment, arrangement and mix module, there are several intermediate outputs, such as ABC notation after the transcription of generated melody, the standard and corresponding output score of multi-agent assessment system, and the arrangement and mix scheme. Those intermediate results are shown in \cref{fig:stage_results}.  

\begin{table*}[ht]
  \centering
  \setlength{\abovecaptionskip}{1.5mm}
  \small 
  \tabcolsep=0.28cm
  \caption{Agents' response accuracy during iterative optimization.}
  \begin{tabular}{l|ccccc} 
  \midrule
   task& organized& +\quad speaking order& +\quad Role-Play& +\quad
Few-Shot Prompting& +\quad Chain of Thought  \\ 
  \midrule
    assessment & 77\% & 80\% &82\% &85\%& 92\%\\
    arrangement & 47\% & 53\% &57\% &64\%& 89\%\\
    mix & 57\% & 67\% &70\% &77\%& 91\%\\
  \bottomrule
  \end{tabular}
\label{tab:agent_ex}
\end{table*}

\subsection{Experiments on Agents}
To validate that the agents function as intended and to elucidate the specific methodologies employed for their improvement and evaluation, we conducted comprehensive experiments. We organize the agents logically and train them by first setting a speaking order and then iteratively incorporating prompt engineering techniques - Role-Play, Few-Shot Prompting and Chain of Thought. During training, the accuracy of agents' responses in assessment, arrangement, and mix tasks improved significantly, reaching around 90\%, shown in \cref{tab:agent_ex}. 

Upon examining the agents' behavior, we found that after setting the speaking order and incorporating Role-Play, the agents never reverse their roles. According to the data, Chain of Thought contributes the most to improving the accuracy of the agents' responses. Among the tasks, the arrangement task exhibits the lowest accuracy before training but demonstrates the most significant improvement after training.

\vspace{5pt}
\begin{figure}[t]
    \centering
    \vspace{-3mm}
    \includegraphics[width=1\linewidth]{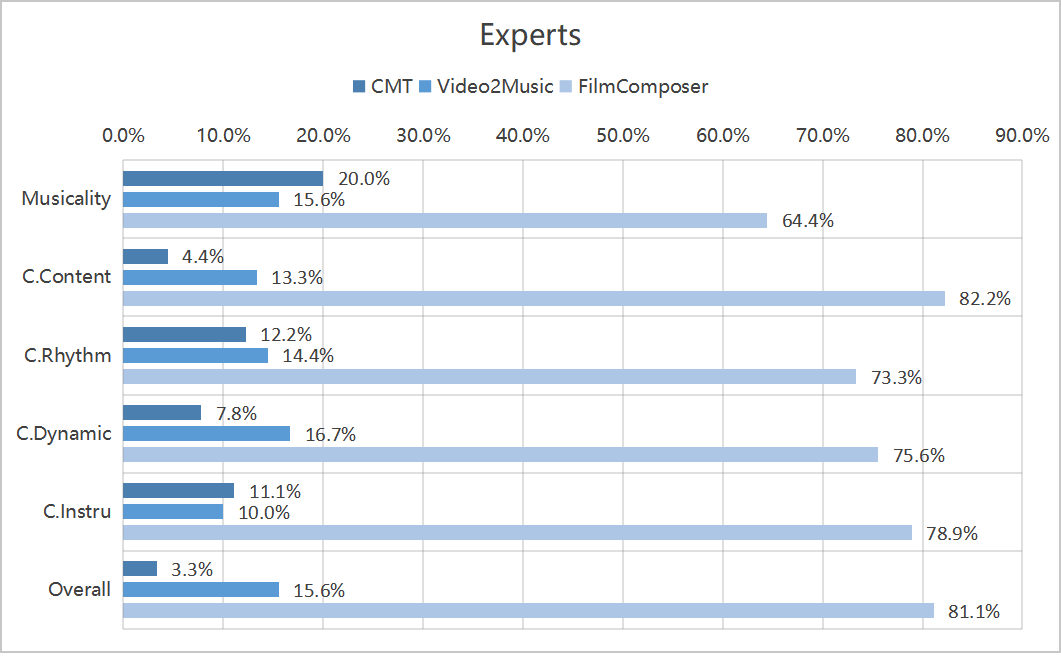}
    \captionof{figure}{The bar chart of the user study of experts, reflecting the scores of the three methods across each item.}
    \label{fig:user-experts}
    \vspace{0mm}
\end{figure}

\vspace{5pt}
\begin{figure}[t]
    \centering
    \vspace{-3mm}
    \includegraphics[width=1\linewidth]{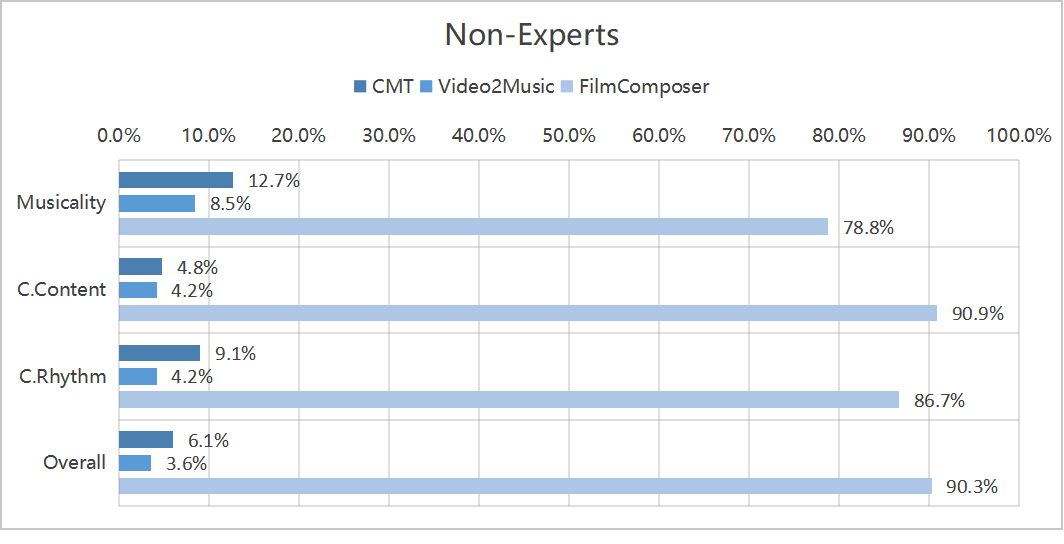}
    \captionof{figure}{The bar chart of the user study of non-experts, reflecting the scores of the three methods across each item.}
    \label{fig:user-non-experts}
    \vspace{0mm}
\end{figure}

\subsection{User Study Details}
In our user study, experts and non-experts were asked to select their preferred results for each item. The results were generated by FilmComposer, CMT, and Video2Music. The scores for each method in each item are presented in \cref{fig:user-experts} and \cref{fig:user-non-experts}.

\subsection{Final Results}
Multiple experimental results and user studies yielded consistent outcomes, demonstrating the strong performance of our method. To further validate this, we provide a video demo and comparison of the results with other models on our project page, showcasing our final results. 

\vspace{5pt}
\begin{figure*}[!t]
    \centering
    \vspace{-3mm}
    \includegraphics[width=1\linewidth]{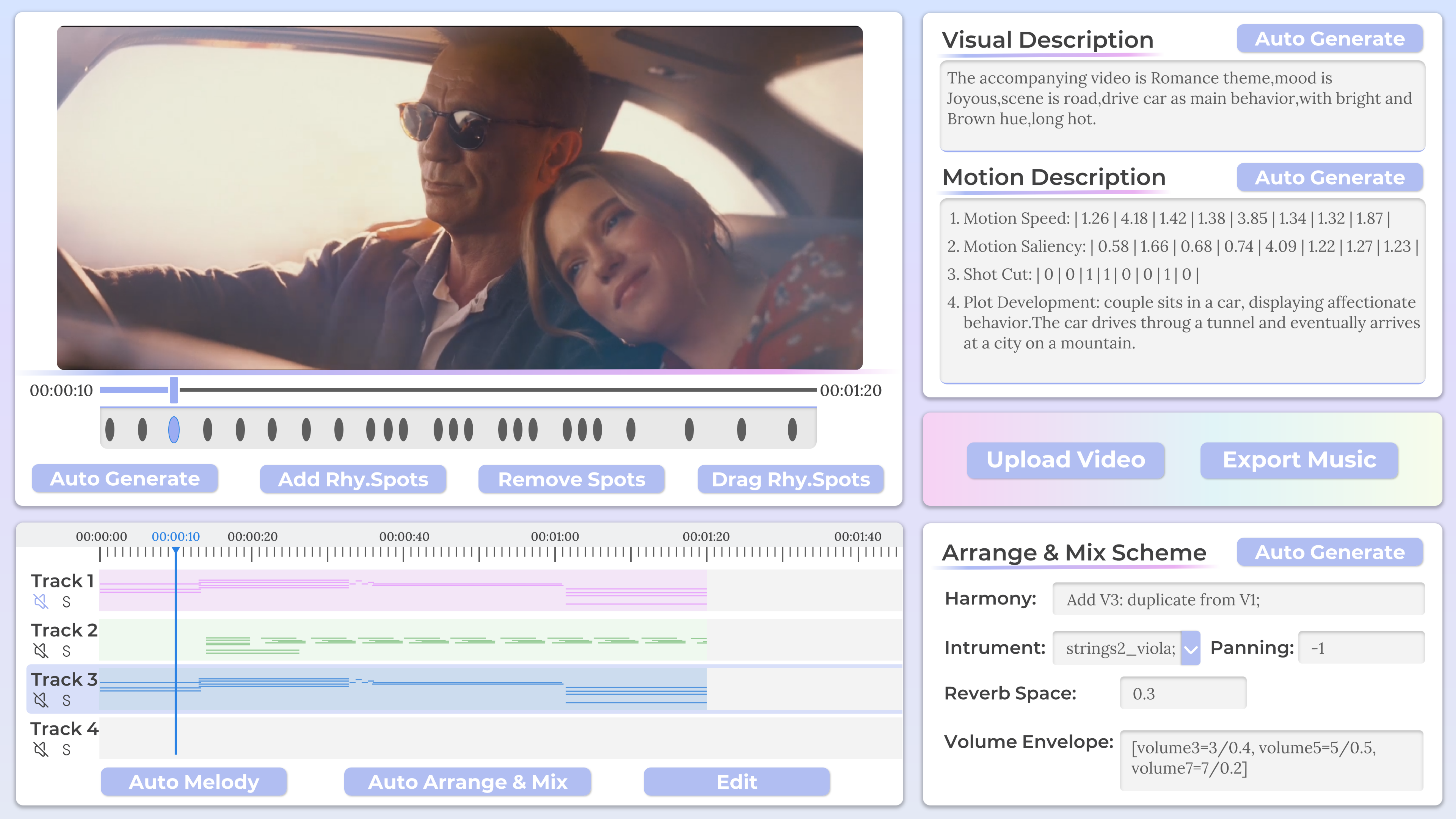}
    \captionof{figure}{The FilmComposer-based interaction system interface.}
    \label{fig:interface}
    \vspace{0mm}
\end{figure*}

\section{Discussion}
\subsection{Analysis of Model Capabilities}
In addition to the information provided by the video, music is also influenced by elements defined by filmmakers, such as mood, genre, and style. Therefore, during the fine-tuning process, we pay particular attention to preserving the model’s ability to process professional descriptions of music as input. By simply incorporating the desired musical elements into the existing visual descriptions, the model can integrate and process these inputs to generate music that aligns with all specified descriptions.

\subsection{Extensions}
The richness and scalability of instruments represent a notable strength of our approach. Our method has covered 39 types of instruments and considered different playing techniques, covering instruments commonly used in film music. Moreover, adding more instruments is easy, as the selection range of Instrument-Agent can easily expand, with numerous sound sources in DAW to be integrated.

Music composition for films sometimes requires consideration of long-term coherence, as the music for a current scene may be influenced by preceding scenes or the storyline. FilmComposer can be extended to effectively capture those long-term dependencies. Module 1 (Visual Processing for Film Clips) can be augmented with a Long Video-Language Understanding model to capture long-term information, which is then fused with the information of current clip via LLMs. This hybrid approach generates effective prompts that simultaneously emphasize localized visual cues and global narrative coherence.

In addition to music, foley constitutes an indispensable auditory component in film productions.  A notable strength of FilmComposer lies in its ability to extend and efficiently generate foley that is precisely synchronized with film clips.  Specifically, this is achieved by incorporating an event timestamp condition, which is analyzed and extracted from the video, thereby enabling the audio generation model to produce foley with accurate temporal alignment to the event.

\subsection{Limitations}
Our extensive instrument repository might lead to potential incompatibility between an instrument's pitch range and melodic requirements. Accurately assessing whether a selected instrument's range fully encompasses the melodic pitch spectrum remains technically demanding. This limitation might result in the omission of notes exceeding the instrument's playable range. 

Future enhancements could address this constraint. Voice-Part-Agent can be introduced to decompose music tracks into distinct voice-part roles, enabling Instrument-Agent to make more informed range-based selections. In addition, implementing algorithmic assistance in instrument selection could establish a dual verification mechanism.

\section{Application Details}
 \cref{fig:interface} shows the interface of the interaction system. The upper left displays the imported video, with tools below allowing for quick rhythm point generation and various adjustments. The upper right provides auto descriptions generation and easy edition. The lower right section is for creating and modifying arrangement and mix plans, and the lower left enables fast music track generation for previewing.

\end{document}